# emLam – a Hungarian Language Modeling baseline

Dávid Márk Nemeskey

Institute for Computer Science and Control
Hungarian Academy of Sciences
`nemeskeyd@gmail.com`

**Abstract.** This paper aims to make up for the lack of documented baselines for Hungarian language modeling. Various approaches are evaluated on three publicly available Hungarian corpora. Perplexity values comparable to models of similar-sized English corpora are reported. A new, freely downloadable Hungarian benchmark corpus is introduced.

## 1 Introduction

Language modeling (LM) is an integral part of several NLP applications, such as speech recognition, optical character recognition and machine translation. It has been shown that the quality of the LM has a significant effect on the performance of these systems [5, 7]. Accordingly, evaluating language modeling techniques is a crucial part of research. For English, a thorough benchmark of n-gram models was carried out by Goodman [10], while more recent papers report results for advanced models [20, 8]. Lately, the One Billion Word Benchmark corpus (1B) [8] was published for the sole reason of measuring progress in statistical language modeling.

The last decade saw dramatic advances in the field of language modeling. Training corpora grew from a few million words (e.g. the Brown corpus) to gigaword, such as 1B, while vocabulary size increased from a few 10k to several hundred thousands. Neural networks [3, 21, 19] overtook n-grams as the language model of choice. State-of-the-art LSTMp networks achieve up to 55% reductions in perplexity compared to 5-gram models [14].

Surprisingly, these developments left few traces in the Hungarian NLP literature. Aside from an interesting line of work on morphological modeling for speech recognition [23, 18], no study is known to the author that addresses issues of Hungarian language modeling. While quality works have been published in related fields, language model performance is often not reported, or is not competitive: e.g. in their otherwise state-of-the-art system, Tarján et al. [28] use a 3-gram model that achieves a perplexity of 400[1] on the test set — a far cry from the numbers reported in [8] and here.

In this paper, we mean to fill this gap in two ways. First, we report baselines for various language modeling methods on three publicly available Hungarian corpora. Hungarian poses a challenge to word-based LM because of its agglutinative nature. The proliferation of word forms inflates the vocabulary and decreases the number of contexts a word form is seen during training, making the data sparsity problem much more pronounced than it is for English. This makes it especially interesting to see how the performance of the tested methods translate to Hungarian.

---
[1] Personal communication with the author.

Second, we present a version of the Hungarian Webcorpus [11] that can be used as a benchmark for LM models. Our motivation was to create the Hungarian equivalent of the One Billion Word Benchmark corpus for English: a freely available data set that is large enough to enable the building of high-quality LMs, yet small enough not to pose a barrier to entry to researchers. We hope that the availability of the corpus will facilitate research into newer and better LM techniques for Hungarian.

The software components required to reproduce this work, as well as the benchmark corpus, comprise the emLam module[2] of e-magyar.hu [30]. The scripts have been released as free software under the MIT license, and can be downloaded from the emLam repository[3].

The rest of the paper is organized as follows. The benchmark corpora, as well as our solution to the data sparsity problem is described in Section 2. In Section 3 we formally define the language modeling task and introduce the methods evaluated. Results are presented in Section 4. Finally, Section 5 contains our conclusions and ideas left for future work.

## 2  The Hungarian Datasets

We selected three publicly available Hungarian corpora for benchmarking. The corpora are of various sizes and domains, which enabled us to evaluate both small- and large-vocabulary LM configurations. The corpus sizes roughly correspond to those of the English corpora commonly used for LM benchmarks, making a comparison between the two languages easier.

The Szeged Treebank [31] is the largest manually annotated corpus of Hungarian. The treebank consists of CoNLL-style tsv files; we used a version in which the morphological features had been converted to KR codes to keep in line with the automatic toolchain described below. At around 1.5 million tokens, it is similar in size to the Penn Treebank [16], allowing us a direct comparison of small-vocabulary LM techniques.

The filtered version of the Hungarian Webcorpus [11] is a semi-gigaword corpus at 589m tokens. It consists of webpages downloaded from the `.hu` domain that contain an "acceptable number of spelling mistakes". The downloadable corpus is already tokenized; we further processed it by performing lemmatization, morphological analysis and disambiguation with Hunmorph [29]: `ocamorph` for the former two and `hunlex` for the latter.

The Hungarian Gigaword Corpus (MNSZ2) [25] is the largest public Hungarian corpus. At around 1G tokens, it is comparable in size to the English 1B corpus. We preprocessed the raw text with the same tools as above.

We decided to use the 'old' `hun*` tools because at the time of writing, the `e-magyar` toolchain was not yet production ready, and the version of the Szeged corpus that uses the new universal POS tags still contained conversion errors. Therefore, the results published here might be slightly different from what one can attain by running the scripts in the `emLam` repository, should the issues above be addressed. However, any such differences will be, most likely, insignificant.

---

[2] http://e-magyar.hu/hu/textmodules/emlam
[3] http://github.com/dlt-rilmta/emLam

## 2.1 Preprocessing

As mentioned before, the main challenge of modeling an agglutinative language is the number of distinct word forms. The solution that works well for English — putting all word forms into the vocabulary — is not reasonable: on one hand, the vocabulary size would explode (see Table 1); on the other, there is a good chance the training set does not contain all possible word forms in the language.

The most common solution in the literature is to break up the words into smaller segments [12, 2, 4]. The two main directions are statistical and morphological word segmentation. While good results have been reported with the former, we opted for the latter: not only is it linguistically more motivated, it also ensures that the tokens we end up with are meaningful, making the LM easier to debug.

We ran the aforementioned pipeline on all words in the corpus, and split all inflectional prefixes (as well as some derivational ones, such as `<COMPAR>`, `<SUPERLAT>`) into separate tokens. Only inflections marked by the KR code are included; the default zero morphemes (the nominative case marker and the present-tense third person singular for verbs) are not. A few examples:

$$\text{jelmondatával} \rightarrow \text{jelmondat <POSS> <CAS<INS>>}$$
$$\text{akartak} \rightarrow \text{akar <PAST> <PLUR>}$$

One could say that by normalizing the text like this, we "deglutenized" it; therefore, the resulting variants of the corpora shall be referred to as "gluten-free" (GLF) from now on.

The full preprocessing pipeline is as follows:

1. Tokenization and normalization. The text was lowercased, converted to `utf-8` and and deglutenized
2. (Webcorpus only) Duplicates sentences were removed, resulting in a 32.5% reduction in corpus size.
3. Tokens below a certain frequency count were converted into `<unk>` tokens. The word distribution proved different from English: with the same threshold as in the 1B corpus (3), much more distinct tokens types remained. To be able to test LMs with a vocabulary size comparable to 1B, we worked with different thresholds for the two gigaword corpora: Webcorpus was cut at 5 words, MNSZ2 at 10. An additional thresholding level was introduced at 30 (50) tokens to make RNN training tractable.
4. Sentence order was randomized
5. The data was divided into train, development and test sets; 90%–5%–5% respectively.

Table 1 lists the main attributes of the datasets created from the three corpora. Where not explicitly marked, the default count threshold (3) is used. The corresponding English corpora are included for comparison. It is clear from comparing the raw and GLF datasets that deglutenization indeed decreases the size of the vocabulary and the number of OOVs by about 50%. Although not shown in the table, this reduction ratio remains consistent among the various thresholding levels.

Also apparent is that, compared to the English corpora, the number of unique tokens is much bigger even in the default Hungarian GLF datasets. Preliminary inquiry into the data revealed that three phenomena account for the majority of the token types between the 3 and 30 (50) count marks: compound nouns, productive derivations and named entities (with mistyped words coming in at fourth place). Since neither the Szeged corpus, nor (consequently) the available morphological disambiguators take compounding and derivation into account, no immediate solution was available for tackling these issues. Therefore, we decided to circumvent the problem by introducing the higher frequency thresholds and concentrating on the problem of inflections in this study.

| Dataset | Sentences | Tokens | Vocabulary | `<unk>`s | Analysis |
|---|---|---|---|---|---|
| Szeged | 81 967 | 1 504 801 | 38 218 | 125 642 | manual |
| Szeged GLF |  | 2 016 972 | 23 776 | 55 067 |  |
| Webcorpus |  | 481 392 824 | 1 971 322 | 5 750 742 | automatic |
| Webcorpus GLF | 26 235 007 | 683 643 265 | 960 588 | 3 519 326 |  |
| Webcorpus GLF-5 |  | " | 625 283 | 4 647 706 |  |
| Webcorpus GLF-30 |  | " | 185 338 | 9 393 015 |  |
| MNSZ2 |  | 624 830 138 | 2 988 629 | 11 614 583 | automatic |
| MNSZ2 GLF | 44 329 309 | 852 232 675 | 1 714 844 | 5 729 509 |  |
| MNSZ2 GLF-10 |  | " | 630 863 | 10 845 301 |  |
| MNSZ2 GLF-50 |  | " | 197 542 | 19 547 859 |  |
| PTB | 49 199 | 1 134 978 | 10 000 |  | manual |
| 1B | 30 607 716 | 829 250 940 | 793 471 |  | automatic |

Table 1. Comparison of the three Hungarian corpora

The preprocessing scripts are available in the emLam repository.

### 2.2  The Benchmark Corpus

Of the three corpora above, the Hungarian Webcorpus is the only one that is freely downloadable and available under a share-alike license (Open Content). Therefore, we decided to make not only the scripts, but the preprocessed corpus as well, similarly available for researchers.

The corpus can be downloaded as a list of tab-separated files. The three columns are the word, lemma and disambiguated morphological features. A unigram (word and lemma) frequency dictionary is also attached, to help create count-thresholded versions. The corpus is available under the Creative Commons Share-alike (CC SA) license.

Such a corpus could facilitate language modeling research in two ways. First, any result published using the corpus is easily reproducible. Second, the fact that it has been preprocessed similarly to the English 1B corpus, makes comparisons such as those in this paper possible and meaningful.

## 3 Language Modeling

The task of (statistical) language modeling is to assign a probability to a word sequence $S = w_1, ..., w_N$. In this paper, we only consider sentences, but other choices (paragraphs, documents, etc.) are also common. Furthermore, we only concern ourselves with *generative* models, where the probability of a word does not depend on subsequent tokens. The probability of $S$ can then be decomposed using the chain rule, as

$$P(S) = P(w_1, ..., w_N) = \sum_{i=1}^{N} P(w_i | w_1, ..., w_{i-1}). \qquad (1)$$

The condition $(w_1, ..., w_{i-1})$ is called the *context* of $w_i$. One of the challenges of language modeling is that the number of possible contexts is infinite, while the training set is not. Because of this, the full context is rarely used; LMs approximate it and deal with the data sparsity problem in various ways.

In the following, we introduce some of the state-of-the-art methods in discrete and continuous language modeling.

### 3.1 N-grams

N-gram models work under the Markov assumption, i.e. the current word only depends on $n-1$ preceding words:

$$P(w_i | w_1, ..., w_{i-1}) \approx P(w_i | w_{i-n+1}, ..., w_{i-1}). \qquad (2)$$

An n-gram model is a collection of such conditional probabilities.

The data sparsity problem is addressed by smoothing the probability estimation in two ways: *backoff* models recursively fall back to coarser $(n-1, n-2, ...$-gram) models when the context of a word was not seen during training, while *interpolated* models always incorporate the lower orders into the probability estimation.

A variety of smoothing models have been proposed over the years; we chose modified Kneser-Ney (KN) [15, 9] as our baseline, since it reportedly outperforms all other n-gram models [10]. We used the implementation in the SRILM [27] library, and tested two configurations: a pruned backoff (the default)[4] and, similar to [8], an unpruned interpolated model[5]. All datasets described in Table 1 were evaluated; in addition, we also tested a GLF POS model, where lemmas were replaced with their respective POS tags.

---

[4] `-kndiscount`
[5] `-kndiscount -gt1min 1 -gt2min 1 -gt3min 1 -gt4min 1 -gt5min 1 -interpolate1 -interpolate2 -interpolate3 -interpolate4 -interpolate5`

### 3.2 Class-based n-grams

Class-based models exploit the fact that certain words are similar to others w.r.t. meaning or syntactic function. By clustering words into classes $C$ according to these features, a class-based n-gram model estimates the probability of the next word as

$$P(w_i|w_1,...,w_{i-1},c_1,...,c_{i-1}) \approx P(w_i|c_i)P(c_{i-n+1},...,c_{i-1}). \qquad (3)$$

This is a Hidden Markov Model (HMM), where the classes are the hidden states and the words are the observations. The techniques proposed for class assignment fall into two categories: statistical clustering [6, 17] and using pre-existing linguistic information such as POS tags [24]. In this paper, we chose the latter, as a full morphological analysis was already available as a by-product of deglutenization.

It is generally agreed that class-based models perform poorly by themselves, but improve word-based models when interpolated with them.

### 3.3 RNN

In the last few years, Recurrent Neural Networks (RNN) have become the mainstream in language modeling research [19, 20, 32, 14]. In particular, LSTM [13] models represent the state-of-the-art on the 1B dataset [14]. The power of RNNs come from two sources: first, words are projected into a continuous vector space, thereby alleviating the sparsity issue; and second, their ability to encode the whole context into their state, thereby "remembering" much further back than n-grams. The downside is that it can take weeks to train an RNN, whereas an n-gram model can be computed in a few hours.

We ran two RNN baselines:

1. the Medium regularized LSTM setup in [32]. We used the implementation[6] in Tensorflow [1]
2. `LSTM-512-512`, the smallest configuration described in [14], which uses LSTMs with a projection layer [26]. The model was reimplemented in Tensorflow, and is available from the emLam repository.

Due to time and resource constraints, the first baseline was only run on the Szeged corpus, and the second only on the smallest, GLF-30 (50) datasets.

### 3.4 Language Model Evaluation

The standard metric of language model quality is *perplexity* (PPL), which measures how well the model predicts the text data. Intuitively, it shows how many options the LM considers for each word; the lower the better. The perplexity of the sequence $w_1,...,w_N$ is computed as

$$PPL = 2^H = 2^{\sum_{i=1}^{N} -\frac{1}{N} \log_2 P(w_i|w_1,...,w_{i-1})}, \qquad (4)$$

where $H$ is the cross-entropy.

---

[6] https://github.com/tensorflow/tensorflow/tree/master/tensorflow/models/rnn/ptb

Language models typically perform worse when tested on a different corpus, due to the differences in vocabulary, word distribution, style, etc. To see how significant this effect is, the models were not only evaluated on the test split of their training corpus, but on the other two corpora as well.

## 4 Evaluation

The results achieved by the n-gram models are reported in Table 2–5. Table 2 lists the perplexities achieved by KN 5-grams of various kinds; the odd one out is POS GLF, where the limited vocabulary enabled us to create up to 9-gram models. For MNSZ2, the reported score is from the 7-gram model, which outperformed 8- and 9-grams.

Similar results reported by others on the PTB and 1B are included for comparison. A glance at the table shows that while word-based 5-grams performed much worse than their counterparts in English, the GLF-based models achieved similar scores.

While the perplexities of GLF models on Webcorpus and MNSZ2 are comparatively close, the perplexities of the word models are about 50% higher on Webcorpus. Finding the cause of this discrepancy requires further research. Two possible candidates are data sparsity (at the same vocabulary size, Webcorpus is 25% smaller) and a difference in the distribution of inflection configurations.

| Corpus | Threshold | Word | GLF | Full POS | POS GLF |
|---|---|---|---|---|---|
| Szeged | 3 | 262.77 | 123.66 | 35.20 | 22.90 |
| Webcorpus | 1 | N/A | N/A | 10.21 | *6.05* |
|  | 5 | 328.22 | 67.90 | N/A | N/A |
|  | 30 | 259.79 | 63.44 | N/A | N/A |
| MNSZ2 | 1 | N/A | N/A | 11.88 | *6.36* |
|  | 10 | 233.52 | 61.92 | N/A | N/A |
|  | 50 | 174.65 | 55.53 | N/A | N/A |
| PTB [22] | N/A | 141.2 |  |  |  |
| 1B [8] | 3 | 90 |  |  |  |

**Table 2.** 5-gram (*9 for POS GLF*) KN test results (PPL)

Table 3 shows the best n-gram perplexities achieved by GLF models. It can be seen that interpolated, unpruned models perform much better than backoff models.

Measuring class-based model performance led to surprising results. As mentioned earlier, the general consensus is that interpolating class- and word-based LMs benefit the performance of the latter; however, our findings (Table 4) did not confirm this. The class-based model could only improve on the unigram model, and failed to do so for the higher orders. The most likely explanation is that as the size of the vocabulary grows larger, the emission entropy increases, which is mirrored by the perplexity. This would explain why class-based n-grams seem to work on small corpora, such as the PTB, but not on MNSZ2.

| Model | pruned backoff | unpruned interpolated |
|---|---:|---:|
| Szeged GLF | 123.66 | **116.32** |
| Webcorpus GLF-5 | 67.90 | **58.62** |
| Webcorpus GLF-30 | 63.44 | **54.42** |
| MNSZ2 GLF-10 | 61.92 | **51.22** |
| MNSZ2 GLF-50 | 55.53 | **46.24** |
| PTB [22] | 141.2 | N/A |
| 1B [8] | 90 | **67.6** |

**Table 3.** The best KN 5-gram results

Another point of interest is the diminishing returns of PPL reductions as the n-gram orders grow. While we have not experimented with 6-grams or higher orders, it seems probable that performance of GLF models would peak at 6- or 7-grams on MNSZ2 (and Webcorpus). For word-based models, this saturation point arrives much earlier: while not reported in the table, the perplexity difference between 4- and 5-gram models is only 1-2 point. This implies that GLF models are less affected by data sparsity.

| Model | GLF-10 | POS $\rightarrow$ GLF-10 | GLF-50 | POS $\rightarrow$ GLF-50 |
|---|---:|---:|---:|---:|
| 1-gram | 2110 | 653.67 | 2175 | 568.53 |
| 2-gram | 127.17 | 327.74 | 115.38 | 285.72 |
| 3-gram | 84.81 | 294.20 | 73.31 | 256.60 |
| 4-gram | 66.06 | 274.70 | 59.41 | 239.64 |
| 5-gram | 61.92 | 261.79 | 55.53 | 228.40 |

**Table 4.** Class (POS)-based model performance on the MNSZ2

It is a well-known fact that the performance of LMs degrade substantially when they are not evaluated on the corpus they were trained on. This effect is clearly visible in Table 5. It is also evident, however, that GLF datasets suffer from this problem to a much lesser extent: while the perplexity more than doubled for the word-based MNSZ2 LMs, it only increased by 50–60% for GLF models. A similar effect can be observed between the full and GLF POS models.

Interestingly, the Webcorpus word models exhibit the smallest perplexity increase of 10-15%. Contrasting this result with Table 2 seems to suggest that there exists a trade-off between predictive power and universality. However, it is worth noting that the performance of these word models still lags well behind that of GLF models.

Finally, Table 6 reports the perplexities achieved by the RNN models. Two conclusions can be drawn from the numbers. First, in line with what has been reported for English by many authors, RNNs clearly outperform even the best n-gram models. Second, the numbers are similar to those reported in the original papers for English. This,

Szeged, 2017. január 26-27.                                                                 9

| Model | Evaluated on | 1 tokens | 5 (10) tokens | 30 (50) tokens |
|---|---|---|---|---|
| Webcorpus word | MNSZ2 | | 377.88 | 291.98 |
| MNSZ2 word | Webcorpus | | 566.60 | 397.13 |
| Webcorpus GLF | MNSZ2 | | 109.71 | 94.59 |
| MNSZ2 GLF | Webcorpus | | 92.51 | 84.91 |
| Webcorpus Full POS | MNSZ2 | 16.14 | | |
| MNSZ2 Full POS | Webcorpus | 16.49 | | |
| Webcorpus POS GLF | MNSZ2 | 8.35 | | |
| MNSZ2 POS GLF | Webcorpus | 7.73 | | |

**Table 5.** Cross-validation results between Webcorpus and MNSZ2 with various thresholds.

together with similar observations above for n-grams, proves that once the "curse of agglutination" is dealt with, a GLF Hungarian is no more difficult to model than English.

| Model | Dataset | Perplexity |
|---|---|---|
| Medium regularized | Szeged GLF | 35.20 |
| LSTM-512-512 | Webcorpus GLF-30 | 40.46 |
| LSTM-512-512 | MNSZ2 GLF-50 | 38.78 |
| Medium regularized [32] | PTB | 82.07 |
| LSTM-512-512 [14] | 1B | 54.1 |

**Table 6.** LSTM model performance

## 5  Conclusion

This work contributes to Hungarian language modeling in two ways. First, we reported state-of-the-art LM baselines for three Hungarian corpora, from million to gigaword size. We found that raw, word-level LMs performed worse than they do for English, but when the text was split into lemmas and inflectional affixes (the "gluten-free" format), results were comparable to those reported on similar-sized English corpora.

Second, we introduced a benchmark corpus for language modeling. To our knowledge, this is the first such dataset for Hungarian. This specially prepared version of the Hungarian Webcorpus is freely available, allowing researchers to easily and reproducibly experiment with new language modeling techniques. It is comparable in size to the One Billion Word Benchmark corpus of English, making comparisons between the two languages easier.

### 5.1 Future Work

While the methods reported here can be called state-of-the-art, many similarly effective modeling approaches are missing. Evaluating them could provide additional insight into how Hungarian "works" or how Hungarian and English should be modeled differently. Understanding the unusual behaviour of word models on Webcorpus also calls for further inquiry into language and corpus structure.

The performance of the models here was measured in isolation. Putting them into use (maybe with some adaptation) in NLP applications such as ASR or ML could answer the question of whether the reduction in perplexity translates to similar reductions in WER or BLEU.

The most glaring problem touched upon, but not addressed, in this paper, is the effect of compounding and derivation on vocabulary size. A way to reduce the number of words could be a more thorough deglutenization algorithm, which would split compound words into their parts and strip productive derivational suffixes, while leaving frozen ones such as `ház·as·ság` untouched. This could indeed be a case when a gluten free diet does make one slimmer.

## Acknowledgements

This work is part of the `e-magyar` framework and was supported by the Research Infrastructure Development Grant, Category 2, 2015 of the Hungarian Academy of Sciences.